\documentclass[10pt,twocolumn,letterpaper]{article}

\usepackage{cvpr}      % To produce the REVIEW version
\usepackage{graphicx}
\usepackage{amsmath}
\usepackage{amssymb}
\usepackage{booktabs}

\usepackage[pagebackref,breaklinks,colorlinks]{hyperref}

\usepackage[capitalize]{cleveref}
\crefname{section}{Sec.}{Secs.}
\Crefname{section}{Section}{Sections}
\Crefname{table}{Table}{Tables}
\crefname{table}{Tab.}{Tabs.}

\def\cvprPaperID{*****} % *** Enter the CVPR Paper ID here
\def\confName{CVPR}
\def\confYear{2022}

\begin{document}

\title{Geospatial Foundational Embedder: Top-1 Winning Solution \\ on EarthVision Embed2Scale Challenge (CVPR 2025)}

\author{Zirui Xu*, Raphael Tang*, Mike Bianco*, Qi Zhang \\
Rishi Madhok, Nikolaos Karianakis, Fuxun Yu$^\dagger$ \\
Microsoft\\
Redmond, WA\\
{\tt\small \{v-ziruixu, v-raptang, v-mikebianco, rishi.madhok, nikarian, fuxunyu\}@microsoft.com}
\thanks{: Equal Contributions. $^\dagger$: Corresponding Author.}
}

\maketitle

%%%%%%%%% ABSTRACT
\begin{abstract}
   EarthVision Embed2Scale challenge (CVPR 2025) aims to develop foundational geospatial models to embed SSL4EO-S12 hyperspectral geospatial data cubes~\cite{ssl4eo} into embedding vectors that faciliatetes various downstream tasks, e.g., classification, regression, etc.
   In this technical report, we introduce our proposed method for the Top-1 winning solution on the Embed2Scale Challenge. 
\end{abstract}

%%%%%%%%% BODY TEXT
\section{Introduction}
\label{sec:intro}

\begin{figure}
    \centering
    \includegraphics[width=0.99\linewidth]{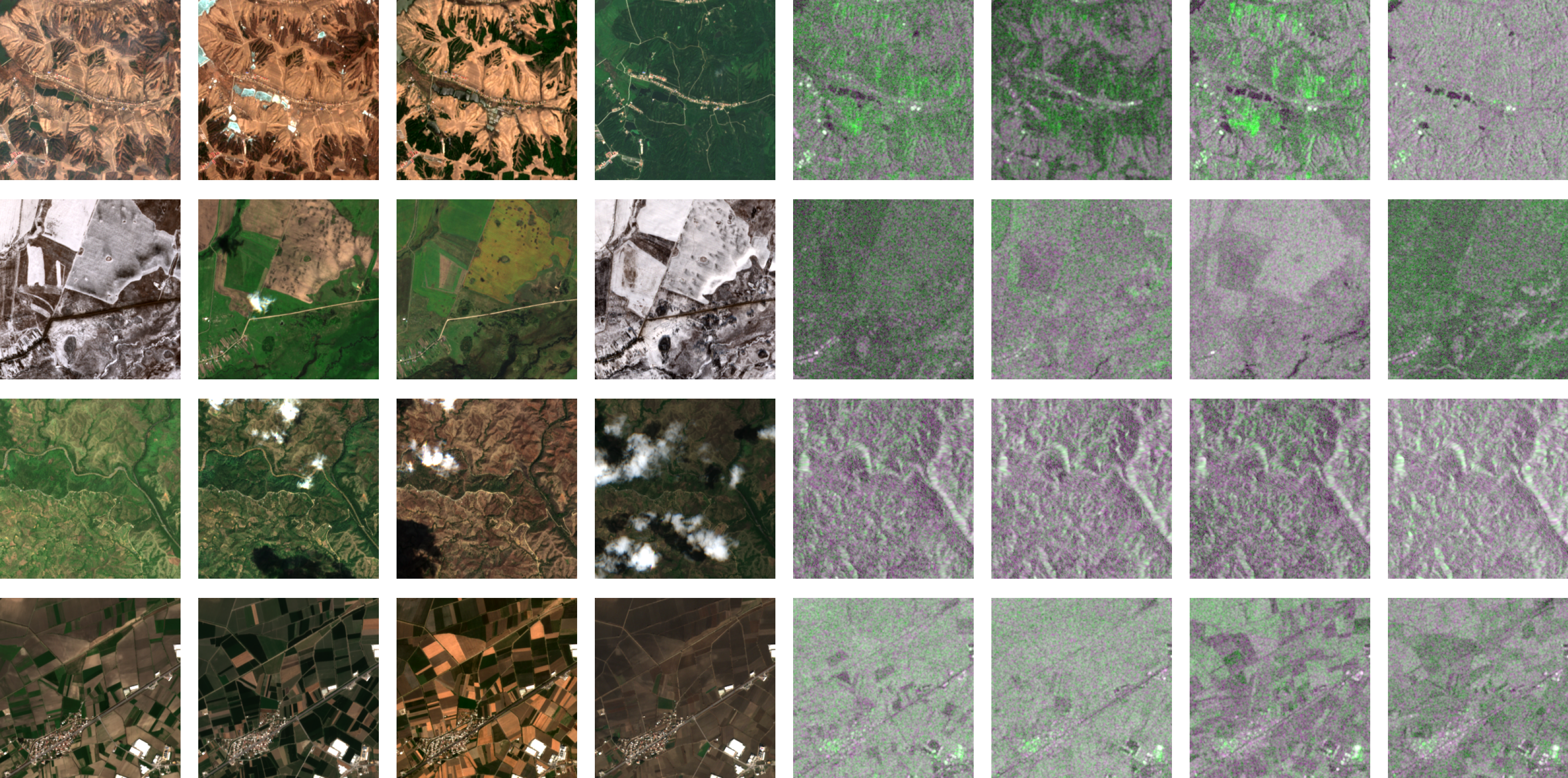}
    \caption{SSL4EO dataset~\cite{ssl4eo}: Each embedding target in the challenge is one novel 4-season hypersepctral geospatial data cube with 4 x 17 = 128 channels.}
    \label{fig:ssl4eo}
\end{figure}

The ability to effectively extract and analyze geospatial information from satellite images is essential for developing solutions that address global challenges including climate change, urbanization, disaster response. Geospatialtial embedding model tackles such tasks by transforming complex geospatial data into meaningful vector representations, facilitating a wide range of applications such as fine-grained scene classification and regression to enable rapid downstream finetuning and response. 
To facilitate that, the EarthVision Embed2Scale competition in CVPR 2025 aimed to advance the field of geospatial embedding by focusing on developing models capable of effectively compressing information from diverse-seasonal, hyperspectral imagery to improve predictive accuracy and robustness in downstream applications, as shown in Figure~\ref{fig:ssl4eo}.

In this competition, we pretrained a series of foundational geospatial models on the given SSL4EO dataset and other external datasets (as we will introduce later). Benefited from the model quality, our embedding quality achieved the 1st place in the various downstream tasks, which underscores the effectiveness of our approach and its potential impact on the field of geospatial learning.

In summary, our main contributions are as follows:
\begin{itemize}
    
    \item State-of-the-Art Pretraining Models: We adopt two families of powerful pretraining methods: CLIP~\cite{clip}, DINO~\cite{dino}. With adapted input channel numbers, we finetuned them on the EO4SSL dataset;

    \item Geospatial Semantic Enrichment: By incorporating multiple auxiliary regression signals~\cite{torchspatial} such as elevation, forest coverage, population and nightlights, we enhance the holistic geospatial understanding and predictive capabilities of our models;
    
    \item Geospatial Natural Language Supervision: Our approach uniquely adopts  natural-language supervised geospatial CLIP~\cite{georsclip}, leveraging their rich semantic strengths to enhance overall embedding performance;
    
    \item SVD-based Compressive Ensemble: We design an SVD + concat method to reduce dimensionality while preserving critical information, and compose the final embeddings through four distinct models, enabling more diverse representation capability.
\end{itemize}

\begin{table*}[!t]
\renewcommand\arraystretch{1.3}
\caption{The Overview of Proposed Embedding Composition Solution. *: same model.}
\centering
\small
\begin{tabular}{lllll}
\hline\hline
\textbf{Embedding} & \textbf{Model Arch}    & \textbf{Input Dimension}        & \textbf{Training Dataset}      & \textbf{Training Method}                           \\ \hline
{[}0:128{]}    & ConvNeXt-XXL & 128-channel          & SSL4EO + TorchSpatial & CLIP w/ lat, lon, four regression targets \\ \hline
{[}128:384{]}  & ViT-Huge     & 128-channel          & SSL4EO                & CLIP w/ lat, lon                          \\ \hline
{[}384:512{]}  & ViT-Base     & 128-channel          & SSL4EO                & DINO                                      \\ \hline
{[}512:640{]}  & ViT-Huge*    & 3-channel RGB [Spring] & RS5M                  & GeoRSCLIP w/ natural language captions                       \\ \hline
{[}640:768{]}  & ViT-Huge*    & 3-channel RGB [Summer] & RS5M                  & GeoRSCLIP w/ natural language captions                       \\ \hline
{[}768:896{]}  & ViT-Huge*    & 3-channel RGB [Fall]   & RS5M                  & GeoRSCLIP w/ natural language captions                       \\ \hline
{[}896:1024{]} & ViT-Huge*    & 3-channel RGB [Winter] & RS5M                  & GeoRSCLIP w/ natural language captions                       \\ \hline\hline
\end{tabular}
\label{table:overview}
\end{table*}

\newpage

\section{The Winning Solution}

\subsection{CLIP and DINO Finetuning}

Contrastive Language-Image Pre-training (CLIP) has revolutionized multimodal learning by enabling the effective alignment of image and text embeddings through contrastive learning objectives. 
% By training on large-scale image-text pairs, CLIP models acquire robust multimodal representations that generalize remarkably well to diverse downstream tasks. The key advantage of CLIP is its ability to embed semantically similar visual and textual concepts closely together in a joint embedding space, facilitating tasks such as cross-modal retrieval and image classification from descriptive labels without explicit task-specific training.
Complementarily, DINO is a powerful self-supervised vision training framework. Unlike CLIP's reliance on paired textual descriptions, DINO utilizes self-distillation principles, encouraging a student network to emulate outputs from a momentum-updated teacher network on augmented views of the same image. This method excels at learning discriminative image embeddings without explicit annotations, resulting in robust representations to capture subtle visual distinctions.

In our solution, we finetuned two powerful foundational model architectures from CLIP off-the-shelf pre-trained weights: ConvNext-XXLarge and ViT-H. We intentionally choose one Convolutional Neural Network (CNN) and one Vision Transformer (ViT) architecture to leverage their inductive bias differences: convolution spatial invariance v.s. transformer's global attention mechanisms. For DINO method, we finetuned a smaller model, ViT-Base, due to the limitation from the off-the-shelf pretrained weights availability. All models' first layer are modified from 3-channel to 128-channel, and we duplicate the 3-channel weights to 128-channel as well to match the new layer shape.

\subsection{Geospatial Semantics Enrichment}

For the two aforementioned methods, DINO is a image-level self-contrasting method and can be hard to fuse further supervision information into training. Therefore, we leave the training setting unmodified for DINO. For CLIP, on the other hand, we considered multiple methods to enrich the geospatial semantics in both pretraining / finetuning stages. 
\begin{itemize}
    \item \textbf{Geospatial Location Semantics}: The training data cubes from EO4SSL dataset comes with $<lat,~lon>$ metadata. We adopt the numerical values and formatted as string: ``\textit{Latitute: lat, Longtitute: lon}" as one part of the supervision to the CLIP language encoder. 
    \item \textbf{Geospatial Regression Semantics}: We also collect four geospatial semantical information for the training data cubes based on their $<lat,~lon>$ metadata. Specifically, we sourced the following four attributes from TorchSpatial~\cite{torchspatial}: Forest Cover, Elevation, Nightlights, and Population Density. And we format the string in the same way: ``\textit{Forest Cover: x1, Elevation: x2, Nightlights: x3, Population: x4}" as the other part of the supervision to the CLIP language encoder. 
    \item \textbf{Natural Language Semantics}: Natural language supervision is hard to get for the given SSL4EO dataset. Therefore, we adopt a natural language geospatial model, GeoRSCLIP~\cite{georsclip}, that is pretrained on the RGB-based RS5M dataset. We use the model as-is on the Sentinel-2 L2A RGB channels of the SSL4EO data cubes for each of the four seasons separately. 
\end{itemize}

\subsection{GeoRSCLIP Unsupervised Finetuning}

During the development phase, we hypothesized that more discriminative, linearly separable features would improve end-task effectiveness because the test-time models were either linear or logistic regression, specifically without a bias term.
Past work has also noted increased linear separability to improve out-of-distribution generalization for graphical embeddings\cite{baranwal2022graph}. 
Thus, we constructed a set of pseudolabels using agglomerative clustering over the four-season concatenated GeoRSCLIP embeddings.
Then, we applied a linear map (i.e., a square matrix) tied between the four seasons' embeddings and fitted a linear model without a bias term (representing the test-time setup) on the subsequent linearly mapped embeddings to predict the pseudolabels.
The intuition was that a full-rank square matrix should distort the embedding space to improve the ``learnability'' for linear models without overfitting to our limited dataset.
Both the linear map and linear model were optimized end-to-end on the development set, and we used the fully optimized linear map to transform and produce our final GeoRSCLIP embeddings.  

\subsection{Compressive Ensemble}

Ensemble distinct models to leverage their different characteristics is the key to enhance the final embedding quality. However, each off-the-shelf model's embedding size are very large: ConvNeXt-XXL / ViT-H has a final 1024 embedding size, and ViT-B has a 768 embedding size, etc. Therefore, we investigate different compressive ensemble methods. Various of methods are evaluated such as post-trained auto-encoder with different MSE / Lp-norm losses, or traditional PCA / SVD algorithms, etc. We empirically found the classic SVD algorithm tends to have less performance impact for the downstream tasks. Therefore, we adopt the SVD+Concat based compressive ensemble method on top of the different models' embeddings.

Table~\ref{table:overview} lists the final models we used to compose the final embedding vector. Specifically, we have four models: 
\begin{itemize}
    \item ConvNeXt-XXL finetuned on SSL4EO dataset by CLIP w/ lat, lon, plus four regression targets from TorchSpatial dataset;
    \item ViT-Huge finetuned on SSL4EO by CLIP w/ lat, lon;
    \item ViT-Base finetuned by DINO on SSL4EO;
    \item ViT-Huge from off-the-shelf natural-language pretraiend GeoRSCLIP.
\end{itemize}
For each model's output embedding, we adopt TruncatedSVD algorithm~\cite{svd} to compress the original embedding size to the target dimension. We utilize multiple numerical evaluation methods to evaluate the pre- and post-compression quality loss. This include MSE reconstruction error, and Kmeans + Silhouette score~\cite{score}. Based on these scores, we search different compression rates for each model and empirically determine the best compression rate combination for different models. The final compressed embedding size for each model is shown in Table~\ref{table:overview}.
\section{Results}

\begin{figure}[!t]
    \centering
    \includegraphics[width=0.8\linewidth]{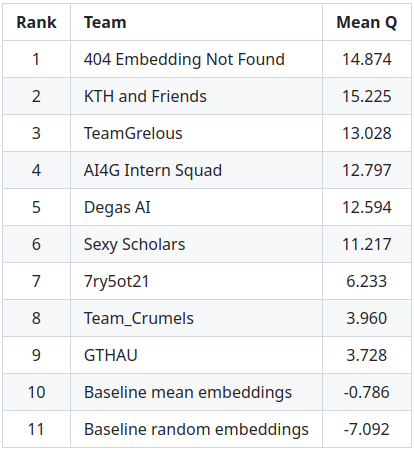}
    \caption{The TEST Phase Leaderboard. Interestingly, although our vanilla unweighted $q\_mean$ score of our model is secondary, we are ranked Top-1 on the task-balanced $q\_mean$ score (calculated from the competition backend, which is used to determine the final rank), denoting stronger generalization capability of our model embedding quality across different tasks.}
    \label{fig:leaderboard}
\end{figure}

\paragraph{Overview} 
Our ensemble embedding method ranks Top-1 in the TEST phase leaderboard, as shown in Figure~\ref{fig:leaderboard}, with the team name: $<$404 Embedding Not Found$>$. 

Note that interestingly, although our vanilla \textbf{unweighted $q\_mean$ score} of our model is secondary, we are ranked Top-1 on the \textbf{task-balanced $q\_mean$ score} for the final leaderboard. In detail~\footnote{Detailed explanations of the final score balancing could be found here: https://eval.ai/web/challenges/challenge-page/2465/evaluation}, the task-balanced weighting mechanism is dependent on the standard deviation of all participants score for that specific task. And a fair weighting is designed by the program who has visibility to all participant teams' performance such that downstream tasks with little difference between the participants' performance have lower weight compared to tasks with a large difference in performance. Such a difference denotes stronger generalization capability of our model embedding quality across different tasks than other teams.

\section{Conclusion}

Geospatialtial embedding helps downstream tasks by transforming complex geospatial pixels into meaningful vector representations, facilitating a wide range of applications such as fine-grained scene classification and regression to enable rapid downstream finetuning and response. In this paper, we introduce our methods of developing and ensembling a series of geospatial foundational models that win the Top-1 prize in the Embed2Scale comptition. We hope the method outlined in this technical report could shine the lights on future geospatial embedding model development.

%%%%%%%%% REFERENCES
{\small
\bibliographystyle{ieee_fullname}
\bibliography{egbib}
}

\end{document}